\documentclass[11pt,a4paper]{article}
\usepackage[hyperref]{acl2019}
\usepackage{times}
\usepackage{latexsym}
\usepackage{booktabs}
\usepackage{multirow,multicol,tabularx}
\usepackage{todonotes}
\usepackage{bbm}
\usepackage{amsmath,bm}

\usepackage{url}

\usepackage{graphicx,accents}

\DeclareMathOperator{\lstm}{LSTM}
\DeclareMathOperator{\softmax}{softmax}

\makeatletter
\DeclareRobustCommand{\cev}[1]{%
  \mathpalette\do@cev{#1}%
}
\newcommand{\do@cev}[2]{%
  \fix@cev{#1}{+}%
  \reflectbox{$\m@th#1\vec{\reflectbox{$\fix@cev{#1}{-}\m@th#1#2\fix@cev{#1}{+}$}}$}%
  \fix@cev{#1}{-}%
}
\newcommand{\fix@cev}[2]{%
  \ifx#1\displaystyle
    \mkern#23mu
  \else
    \ifx#1\textstyle
      \mkern#23mu
    \else
      \ifx#1\scriptstyle
        \mkern#22mu
      \else
        \mkern#22mu
      \fi
    \fi
  \fi
}

\makeatother

\aclfinalcopy % Uncomment this line for the final submission
\title{Improving Interaction Quality Estimation with BiLSTMs and the\\ Impact on Dialogue Policy Learning}

\author{Stefan Ultes \\
  Daimler AG \\
  Sindelfingen, Germany \\
  \texttt{stefan.ultes@daimler.com}}

\date{}

\begin{document}
\maketitle
\begin{abstract}
Learning suitable and well-performing dialogue behaviour in statistical spoken dialogue systems has been in the focus of research for many years. While most work which is based on reinforcement learning employs an objective measure like task success for modelling the reward signal, we use a reward based on user satisfaction estimation. We propose a novel estimator and show that it outperforms all previous estimators while learning temporal dependencies implicitly. Furthermore, we apply this novel user satisfaction estimation model live in simulated experiments where the satisfaction estimation model is trained on one domain and applied in many other domains which cover a similar task.
We show that applying this model results in higher estimated satisfaction, similar task success rates and a higher robustness to noise.
\end{abstract}

\section{Introduction}
% Spoken dialogue systems (SDSs) enable voice interaction between technical systems and humans. 
One prominent way of modelling the decision-making component of a spoken dialogue system (SDS) is to use (partially observable) Markov decision processes ((PO)MDPs)~\cite{lemon2012,young2013}. There, reinforcement learning (RL)~\cite{sutton1998} is applied to find the optimal system behaviour represented by the policy $\pi$. Task-oriented dialogue systems model the reward $r$, used to guide the learning process, traditionally with task success as the principal reward component~\cite{gasic2014gaussian,lemon2007machine,daubigney2012,levin1997,young2013,su2015,su2016acl}.

An alternative approach proposes user satisfaction as the main reward component~\cite{ultes2017domain}. However, the applied statistical user satisfaction estimator heavily relies on handcrafted temporal features. Furthermore, the impact of the estimation performance on the resulting dialogue policy remains unclear.

In this work, we propose a novel LSTM-based user satisfaction reward estimator that is able to learn the temporal dependencies implicitly and compare the performance of the resulting dialogue policy with the initially used estimator. 
%The contribution is two-fold: 
% \begin{itemize}
%     \item We propose a novel deep learning-based user satisfaction estimator outperforming all previous estimators for the same task.
%     \item We analyse the effects of two user satisfaction estimators on the performance of the learned dialogue behaviour.
% \end{itemize}

Optimising the dialogue behaviour to increase user satisfaction instead of task success has multiple advantages:
\begin{enumerate}
  \item The user satisfaction is more domain-independent as it can be linked to interaction phenomena independent of the underlying task~\cite{ultes2017domain}.
  \item User satisfaction is favourable over task success as it represents more accurately the user's view and thus whether the user is likely to use the system again in the future. Task success has only been used as it has shown to correlate well with user satisfaction~\cite{williams2004characterizing}.
\end{enumerate}

% In this contribution, the emphasis lies on the second and third item. %(we leave the last item for future work). 
% Following up on previous work~\cite{ultes2011b,ultes2012,ultes2014,ultes2015b,ultes2017},  
Based on previous work by~\citet{ultes2017domain}, the
interaction quality (IQ)---a less subjective version of user satisfaction\footnote{The relation of US and IQ has been closely investigated in \cite{schmitt2015,ultes2013a}.}---will be used for estimating the reward. The estimation model is thus based on domain-independent, interaction-related features which do not have any information available about the goal of the dialogue. This allows the reward estimator to be applicable for learning in unseen domains.

The originally applied IQ estimator heavily relies on handcrafted temporal features. In this work, we will present a deep learning-based IQ estimator that utilises the capabilities of recurrent neural networks to get rid of all handcrafted features that encode temporal effects. By that, these temporal dependencies may be learned instead.

\begin{figure*}[t]
  \centering
  \includegraphics[width=0.75\linewidth]{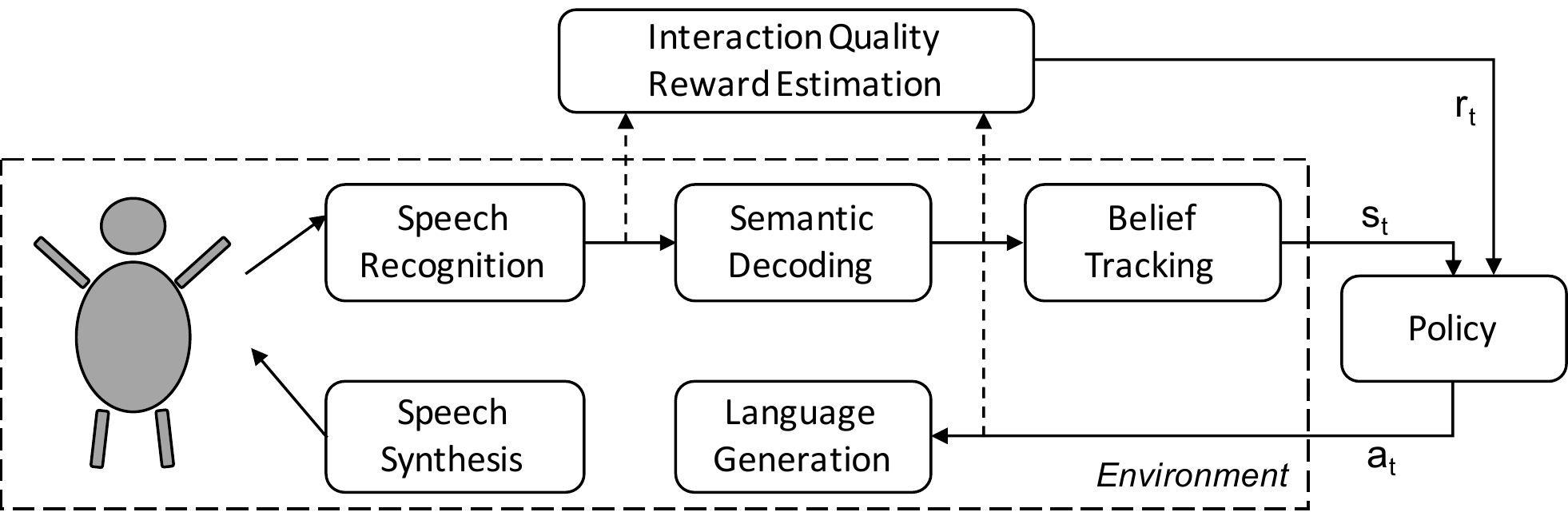}
  \caption{The RL framework integrating an interaction quality reward estimator as proposed by \citet{ultes2017domain}. The policy learns to take action $a_t$ at time $t$ while being in state $s_t$ and receiving reward $r_t$.%\vspace{-14pt}
  }
  \label{fig:RLframework}
\end{figure*}

The applied RL framework is shown in Figure~\ref{fig:RLframework}. Within this setup, both IQ estimators are used for learning dialogue policies in several domains to analyse their impact on general dialogue performance metrics.

%The model has been trained on manually annotated dialogue turns of a bus information system achieving an accuracy of 0.89\footnote{taking into account neighbouring values, cf.~Sec.~\ref{par:iqestimator}}.
%\mgcomment{Although you cite your previous work here, I think it would be helpful if you actually say how you train the user satisfaction model, what kind of data you need, and how accurate it is.} 
% The applied RL framework is shown in Figure~\ref{fig:RLframework}. It has previously been applied for in-domain experiments and simulated evaluation~\cite{ultes2017}. In this paper, we will complete the work by using a modified reward function to show its domain-independence and the resulting high potential to be applicable for learning in unseen domains. Moreover, the estimator is used in an experiment where the policy is learned through interaction with real humans.

The remainder of the paper is organised as follows: in Section~\ref{sec:related_work}, related work is presented focusing on dialogue learning and the type of reward that is applied. In Section~\ref{sec:iq_reward_estimation}, the interaction quality is presented and how it is used in the reward model. The deep learning-based interaction quality estimator proposed in this work is then described in detail in Section~\ref{sec:lstm_estimator} followed by the experiments and results both of the estimator itself and the resulting dialogue policies in Section~\ref{sec:results}.

\section{Relevant Related Work}
\label{sec:related_work}

Most of  previous work on dialogue policy learning focuses on employing task success as the main reward signal~\cite{gasic2014gaussian,gasic2014,lemon2007machine,daubigney2012,levin1997,young2013,su2015,su2016acl}.  However, task success is usually  only computable for predefined tasks e.g., through interactions with simulated or recruited users, where the underlying goal is known in advance. %In real-world scenarios, this is hardly possible. 
To overcome this, the required information can be requested directly from  users at the end of each dialogue~\cite{gasic2013}. However, this can be intrusive, and users may not always cooperate.
%requires a certain degree of willingness to collaborate from the user side which is not always the case. 

An alternative is to use a task success estimator~\cite{elasri2014task, su2015,su2016acl}.  With the right choice of features, these can also be applied to new and unseen domains~\cite{vandyke2015}. %Such a task success estimator is based on the internal representation of the goal of the dialogue. 
However,  these models still attempt to estimate completion of the underlying task, whereas our model evaluates the overall user experience.

In this paper, we show that an interaction quality reward estimator trained on dialogues from a bus information system will result in well-performing dialogues both in terms of success rate and user satisfaction on five other domains, while only using interaction-related, domain-independent information, i.e., not knowing anything about the task of the domain.

Others have previously introduced user satisfaction into the reward~\cite{walker1998learning,walker2000,rieser2008,rieser08} by using the PARADISE framework~\cite{walker1997}. However, PARADISE relies on the existence of explicit task success information which is usually hard to obtain. 

Furthermore, to derive user ratings within that framework, users have to answer a questionnaire which is usually not feasible in real world settings. To overcome this, PARADISE has been used in conjunction with expert judges instead~\cite{elasri2012reward,elasri2013reward} to enable unintrusive acquisition of  dialogues. However, the problem of mapping the results of the questionnaire to a scalar reward value still exists. 

Therefore, we  use  interaction quality (Section~\ref{sec:iq_reward_estimation}) in this work because it uses scalar values applied by experts and only uses task-independent features that are easy to derive. 

% The remainder of the paper is organized as follows: in Section~\ref{sec:iq_reward_estimation}, the interaction quality reward estimation module is presented in detail. Section~\ref{sec:experiments} contains the simulated experiments on several domains as well as an experiment with paid subjects. The findings are discussed in Section~\ref{sec:discussion} and conclusions are drawn in Section~\ref{sec:conclusion}.

\section{Interaction Quality Reward Estimation}
\label{sec:iq_reward_estimation}
In this work, the reward estimator is based on the interaction quality (IQ)~\cite{schmitt2015} for learning information-seeking dialogue policies. IQ represents a less subjective variant of user satisfaction: instead of being acquired from users directly, experts annotate pre-recorded dialogues to avoid the large variance that is often encountered when users rate their dialogues directly~\cite{schmitt2015}. 

IQ is defined on a five-point scale from five (satisfied) down to one (extremely unsatisfied). To derive a reward from this value, the equation 
\begin{equation}
R_{IQ} = T \cdot (-1) + (iq - 1) \cdot 5
\end{equation}
is used where $R_{IQ}$ describes the final reward. It is applied to the final turn of the dialogue of length $T$ with a final IQ value of $iq$. 
A per-turn penalty of $-1$ is added to the dialogue outcome. This results in a reward range of 19 down to $-T$ which is consistent with related work~\cite[e.g.]{gasic2014gaussian,vandyke2015,su2016acl} in which binary task success (TS) was used to define the reward as:
\begin{equation}
R_{TS} = T \cdot (-1) + \mathbbm{1}_{TS} \cdot 20 \; ,
\end{equation}
where $\mathbbm{1}_{TS} = 1$ only if the dialogue was successful, $\mathbbm{1}_{TS} = 0$ otherwise. $R_{TS}$ will be used as a baseline.

\begin{figure}[tp]
\begin{center} 
  \includegraphics[width=0.9\linewidth]{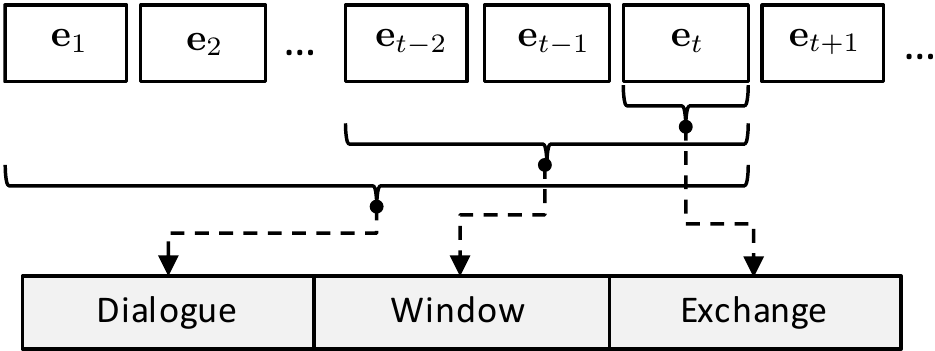}
  \caption{Modelling of temporal information in the interaction parameters used as input to the IQ estimator.%\vspace{-14pt}%\mgcomment{Maybe I missed this but you don't explain this diagram in text}
  }
  \label{fig:parameterlevels}
\end{center}
\end{figure}

% \paragraph %{Interaction Quality Estimation}
\label{par:iqestimator}
The problem of estimating IQ has been cast as a classification problem where the target classes are the distinct IQ values. The input consists of domain-independent variables called interaction parameters. These parameters incorporate information from the automatic speech recognition (ASR) output and the preceding system action.
%: the ASR status (one of \textit{success}, \textit{no match}, \textit{no input}), the ASR confidence of the highest ranked result, the general type of the system action (one of \textit{statement}, \textit{question}), whether the system action is a repeat of the previous system action, and whether the role of the system action is to confirm previous user input. 
Most previous approaches used this information, which is available at every turn, to compute temporal features by taking sums, means or counts from the turn-based information for a window of the last 3 system-user-exchanges\footnote{a system turn followed by a user turn} and the complete dialogue (see Fig.~\ref{fig:parameterlevels}). 
The baseline IQ estimation approach as applied by \citet{ultes2017domain} (and originating from \citet{ultes2015b}) used a feature set of 16 parameters as shown in Table~\ref{tab:parameters} with a support vector machine (SVM)~\cite{vapnik1995,chang2011}.%\footnote{The parameters are ASRRecognitionStatus, ASRConfidence, RePrompt?, \#Exchanges, ActivityType, Confirmation?, MeanASRConfidence, \#ASRSuccess, \%ASRSuccess, \#ASRRejections, \%ASRRejections, \{Mean\}ASRConfidence, \{\#\}ASRSuccess, \{\#\}ASRRejections, \{\#\}RePrompts, \{\#\}SystemQuestions. \{$\cdot$\} denotes the window level.}. 

The LEGO corpus~\cite{schmitt2012a} provides data for training and testing and consists of 200 dialogues (4,885 turns) from the Let's Go bus information system~\cite{raux2006}. 
% of Carnegie Mellon University in Pittsburgh, PA. 
There, users with real needs were able to call the system to get information about the bus schedule. Each turn of these 200 dialogues has been annotated with IQ (representing the quality of the dialogue up to the current turn) by three experts. The final IQ label has been assigned using the median of the three individual labels.

% Table generated by Excel2LaTeX from sheet 'Sheet2'
\begin{table}[t]
  \centering
  \setlength{\tabcolsep}{3pt}
  \footnotesize
  \caption{The parameters used for IQ estimation extracted on the exchange level from each user input plus counts, sums and rates for the whole dialogue (\#,\%,Mean) and for a window of the last 3 turns (\{$\cdot$\}).}
    \begin{tabularx}{\linewidth}{crX}
    \toprule
     & Parameter & Description \\
    \midrule
    \parbox[t]{2mm}{\multirow{8}{*}{\rotatebox[origin=c]{90}{\textit{Exchange level}}}} & ASRRecognitionStatus & ASR status: \textit{success}, \textit{no match}, \textit{no input} \\
    & ASRConfidence & confidence of top ASR results \\
    & RePrompt? & is the system question the same as in the previous turn? \\
    & ActivityType & general type of system action: \textit{statement}, \textit{question} \\
    & Confirmation? & is system action confirm? \\
    \cmidrule{2-3}
    \parbox[t]{2mm}{\multirow{7}{*}{\rotatebox[origin=c]{90}{{\textit{Dialogue level}}}}}& MeanASRConfidence & mean ASR confidence if ASR is success \\
    & \#Exchanges & number of exchanges (turns) \\
    & \#ASRSuccess & count of ASR status is success \\
    & \%ASRSuccess & rate of ASR status is success \\
    & \#ASRRejections & count of ASR status is reject \\
    & \%ASRRejections & rate of ASR status is reject \\
    \cmidrule{2-3}
    \parbox[t]{2mm}{\multirow{8}{*}{\rotatebox[origin=c]{90}{{\textit{Window level}}}}}& \{Mean\}ASRConfidence  & mean ASR confidence if ASR is success \\
    & \{\#\}ASRSuccess & count of ASR is success \\
    & \{\#\}ASRRejections & count of ASR status is reject \\
    & \{\#\}RePrompts & count of times RePromt?\ is true \\
    & \{\#\}SystemQuestions & count of ActivityType is question \\
    \bottomrule
    \end{tabularx}%
  \label{tab:parameters}%
  %\vspace{-12pt}
\end{table}%

Previous work has used the LEGO corpus with a full IQ feature set (which includes additional partly domain-related information) achieving an unweighted average recall\footnote{UAR is the arithmetic average of all class-wise recalls.} (UAR) of 0.55 using ordinal regression~\cite{elasri2014b}, 0.53 using a two-level SVM approach~\cite{ultes2013d}, and 0.51 using a hybrid-HMM~\cite{ultes2014b}. Human performance on the same task is 0.69 UAR~\cite{schmitt2015}. A deep learning approach using only non-temporal features achieved an UAR of 0.55~\cite{rach2017interaction}.

%  achieving an unweighted average recall\footnote{UAR is the arithmetic average of all class-wise recalls.} (UAR) of 0.55 with 10-fold cross-validation. However, missing the correct estimated IQ value by only one has little impact for modelling the reward, and if neighbouring values are taken into account, the model achieves an accuracy of 0.89. %\mgcomment{you should give some insight into what features are used.}
% 
% As a comparison,

\section{LSTM-based Interaction Quality Estimation}
\label{sec:lstm_estimator}

\begin{figure}[t]
\begin{center} 
  \includegraphics[width=0.9\linewidth]{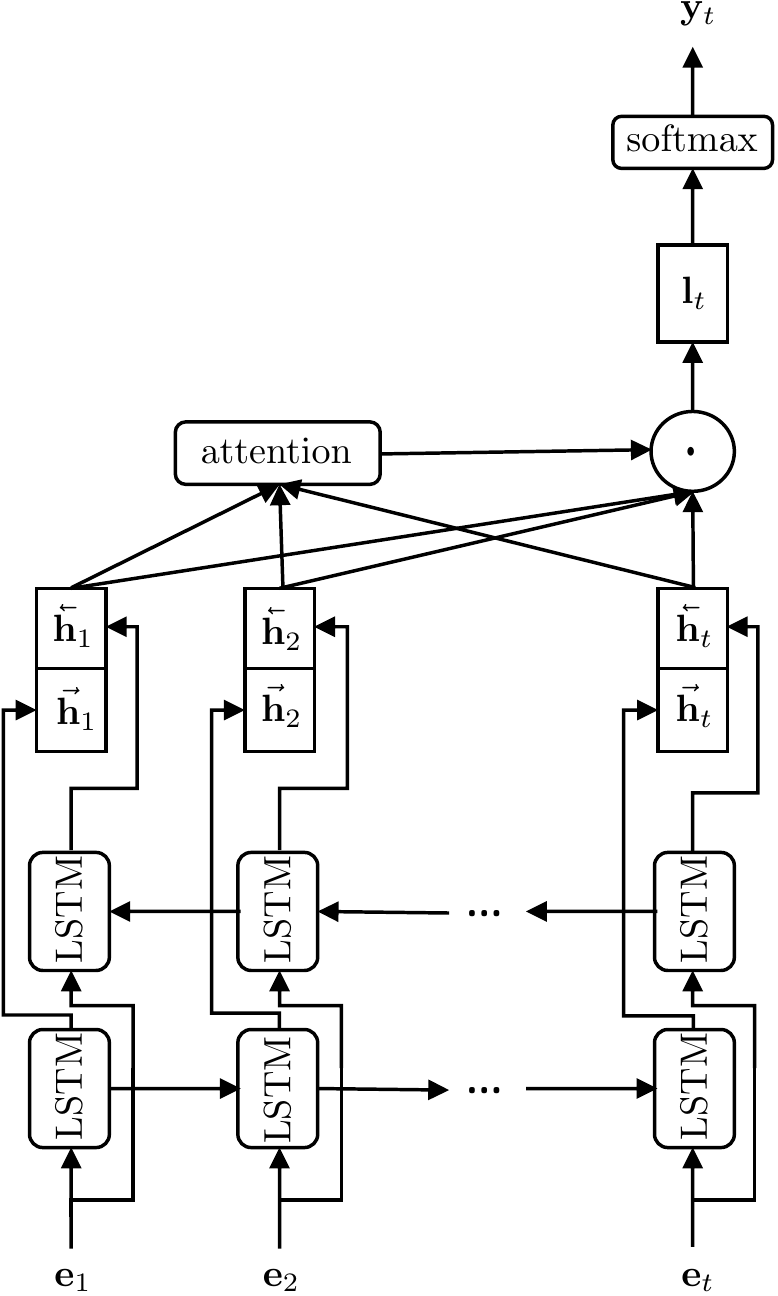}
  \caption{The architecture of the proposed BiLSTM model with self attention. For each time $t$, the exchange level parameter of all exchanges $\mathbf{e}_i$ of the sub-dialogue $i \in \{1 \ldots t\}$ are encoded to their respective hidden representation $\mathbf{h}_i$ and are considered and weighted with the self attention mechanism to finally estimate the IQ value $\mathbf{y}_t$ at time $t$.
  %\vspace{-14pt}
  }
  \label{fig:architecture}
\end{center}
\end{figure}

The proposed IQ estimation model will be used as a reward estimator as depicted in Figure~\ref{fig:RLframework}. With parameters that are collected from the dialogue system modules for each time step $t$, the reward estimator derives the reward $r_t$ that is used for learning the dialogue policy $\pi$.

The architecture of our proposed IQ estimation model is shown in Figure~\ref{fig:architecture}. It is based on the idea that the temporal information that has previously been explicitly encoded with the window and dialogue interaction parameter levels may be learned instead by using recurrent neural networks. Thus, only the exchange level parameters $\mathbf{e}_t$ are considered (see Table~\ref{tab:parameters}). Long Short-Term Memory (LSTM) cells are at the core of the model and have originally been proposed by~\citet{hochreiter1997} as a recurrent variant that remedies the vanishing gradient problem~\cite{bengio1994learning}. 

As shown in Figue~\ref{fig:architecture}, the exchange level parameters form the input vector $\mathbf{e}_t$ for each time step or turn $t$ to a bi-directional LSTM~\cite{graves2013hybrid} layer. The input vector $\mathbf{e}_t$ encodes the nominal parameters ASRRecognitionStatus, ActivityType, and Confirmation? as 1-hot representations. In the BiLSTM layer, two hidden states are computed: $\vec{\mathbf{h}}_t$ constitutes the forward pass through the current sub-dialogue and $\cev{\mathbf{h}}_t$ the backwards pass: 
\begin{align}
\vec{\mathbf{h}}_t &= \lstm(\mathbf{e}_t,\vec{\mathbf{h}}_{t-1}) \\
\cev{\mathbf{h}}_t &= \lstm(\mathbf{e}_t,\cev{\mathbf{h}}_{t+1})
\end{align}
The final hidden layer is then computed by concatenating both hidden states: 
\begin{equation}
\mathbf{h}_t = [\vec{\mathbf{h}}_t , \cev{\mathbf{h}}_t] \; .
\end{equation}

Even though information from all time steps may contribute to the final IQ value, not all time steps may be equally important. Thus, an attention mechanism~\cite{vaswani2017attention} is used that evaluates the importance of each time step $t'$ for estimating the IQ value at time $t$ by calculating a weight vector $\alpha_{t,t'}$. 
\begin{align}
\mathbf{g}_{t,t'} &= \tanh(\mathbf{h}_t^T \mathbf{W}_t + \mathbf{h}_{t'}^T \mathbf{W}_{t'} + \mathbf{b}_t) \\
\bm{\alpha}_{t,t'} &= \softmax(\sigma(\mathbf{W}_a \mathbf{g}_{t,t'} + \mathbf{b}_a)) \\
\mathbf{l}_t &= \sum_{t'}\bm{\alpha}_{t,t'} \mathbf{h}_{t'} 
\end{align}
\citet{zheng2018opentag} describe this as follows: ``The attention-focused hidden state representation $\mathbf{l}_t$ of an [exchange] at time step $t$ is given by the weighted summation of the hidden state representation $\mathbf{h}_{t'}$ of all [exchanges] at time steps $t'$, and their similarity $\bm{\alpha}_{t,t'}$ to the hidden state representation $\mathbf{h}_t$ of the current [exchange]. Essentially, $\mathbf{l}_t$ dictates how much to attend to an [exchange] at any time step conditioned on their neighbourhood context.''

To calculate the final estimate $\mathbf{y}_t$ of the current IQ value at time $t$, a softmax layer is introduced:
\begin{equation}
\mathbf{y}_t = \softmax(\mathbf{l}_t)
\end{equation}

For estimating the interaction quality using a BiLSTM, the proposed architecture frames the task as a classification problem where each sequence is labelled with one IQ value. Thus, for each time step $t$, the IQ value needs to be estimated for the corresponding sub-dialogue consisting of all exchanges from the beginning up to $t$. Framing the problem like this is necessary to allow for the application of a BiLSTM-approach and still be able to only use information that would be present at the current time step $t$ in an ongoing dialogue interaction.

To analyse the influence of the BiLSTM, a model with a single forward-LSTM layer is also investigated where
\begin{equation}
\mathbf{h}_t = \vec{\mathbf{h}}_t \; .
\end{equation}

Similarly, a model without attention is also analysed where
\begin{equation}
\mathbf{l}_t = \mathbf{h}_t \; .
\end{equation}

\section{Experiments and Results}
\label{sec:results}
The proposed BiLSTM IQ estimator is both trained and evaluated on the LEGO corpus and applied within the IQ reward estimation framework (Fig.~\ref{fig:RLframework}) on several domains within a simulated environment.

\subsection{Interaction Quality Estimation}
% \todo[inline]{
% - setup: 
%   - models comapred: 4 proposed, 2 baseline
%   - 10-fold x-cal, call-wise x-val, lego, limit of 60 dialogues
%   - metrics for classification: uar, spearman, weighted cohen, ext acc
%   - keras, self-attention (citation?), categorical crossentropy, rmsprop, minibatches: 16, $lr=0.001$, max epochs: 100, select best epoch based on x-val-result
%   - 
% - results:
%   - traditional x-val pointless, call-wise used
%   - proposed bi with att performs best on unseen data
%   - conclusion: used for dialogue policy learning
% }

% Table generated by Excel2LaTeX from sheet 'LEGO'
\begin{table}[t]
  \centering \footnotesize
  \setlength{\tabcolsep}{4pt}
  \caption{Performance of the proposed LSTM-based variants with the traditional cross-validation setup. Due to overlapping sub-dialogues in the train and test sets, the performance of the LSTM-based models achieve unrealistically high performance.}
    \begin{tabular}{lccccc}
    \toprule
          & \textit{UAR} & $\kappa$ & $\rho$ & \textit{eA} & \textit{Ep.} \\
          \midrule
    LSTM  & 0.78  & 0.85  & 0.91  & \textbf{0.99} & 101 \\
    BiLSTM & \textbf{0.78} & \textbf{0.85} & \textbf{0.92} & \textbf{0.99} & 100 \\
    LSTM+att & 0.74  & 0.82  & 0.91  & \textbf{0.99} & 101 \\
    BiLSTM+att & 0.75  & 0.83  & 0.91  & \textbf{0.99} & 93 \\
    \cmidrule{2-6}
    %\citet{rach2017interaction} & 63.21 & 77.64 & 89.87 & 0.98  & 99 \\
    \citet{rach2017interaction} & 0.55 & 0.68 & 0.83 & 0.94 & - \\
%     \citet{ultes2015b} & 52.14 & 60.53 & 74.25 & 0.89  & - \\
    \citet{ultes2015b} & 0.55 & - & - & 0.89  & - \\
    \bottomrule
    \end{tabular}%
  \label{tab:xvalresults}%
\end{table}%

To evaluate the proposed BiLSTM model with attention (BiLSTM+att), it is compared with three of its own variants: a BiLSTM without attention (BiLSTM) as well as a single forward-LSTM layer with attention (LSTM+att) and without attention (LSTM). Additional baselines are defined by \citet{rach2017interaction} who already proposed an LSTM-based architecture that only uses non-temporal features, and the SVM-based estimation model as originally used for reward estimation by \citet{ultes2015b}.

The deep neural net models have been implemented with Keras~\cite{chollet2015keras} using the self-attention implementation as provided by \citet{zheng2018opentag}\footnote{Code freely available at \url{https://github.com/CyberZHG/keras-self-attention}}. All models were trained against cross-entropy loss using RmsProp~\cite{tieleman2012rmsprop} optimisation with a learning rate of 0.001 and a mini-batch size of 16.

As evaluation measures, the unweighted average recall (UAR)---the arithmetic average of all class-wise recalls---, a linearly weighted version of Cohen's $\kappa$, and Spearman's $\rho$ are used. As missing the correct estimated IQ value by only one has little impact for modelling the reward, a measure we call the extended accuracy (eA) is used where neighbouring values are taken into account as well.

% Table generated by Excel2LaTeX from sheet 'LEGO'
\begin{table}[t]
  \centering \footnotesize
  \setlength{\tabcolsep}{4pt}
  \caption{Performance of the proposed LSTM-based variants with the dialogue-wise cross-validation setup. The models by \citet{rach2017interaction} and \citet{ultes2015b} have been re-implemented. The BiLSTM with attention mechanism performs best in all evaluation metrics.}
    \begin{tabular}{lccccc}
    \toprule
          & \textit{UAR} & $\kappa$ & $\rho$ & \textit{eA} & \textit{Ep.} \\
         \midrule
    LSTM  & 0.51  & 0.63  & 0.78  & 0.93  & 8 \\
    BiLSTM & 0.53  & 0.63  & 0.78  & 0.93  & 8 \\
    LSTM+att & 0.52  & 0.63  & 0.79  & 0.92  & 40 \\
    BiLSTM+att & \textbf{0.54} & \textbf{0.65} & \textbf{0.81} & \textbf{0.94} & 40 \\
    \cmidrule{2-6}
    \citet{rach2017interaction} & 0.45  & 0.58  & 0.79  & 0.88  & 82 \\
    \citet{ultes2015b}   & 0.44  & 0.53  & 0.69  & 0.86  & - \\
    \bottomrule
    \end{tabular}%
  \label{tab:callxvalresults}%
\end{table}%

All experiments were conducted with the LEGO corpus~\cite{schmitt2012a} in a 10-fold cross-validation setup for a total of 100 epochs per fold. The results are presented in Table~\ref{tab:xvalresults}. Due to the way the task is framed (one label for each sub-dialogue), memorising effects may be observed with the traditional cross-validation setup that has been used in previous work. Hence, the results in Table~\ref{tab:xvalresults} show very high performance, which is likely to further increase with ongoing training. However, the corresponding models are likely to generalise poorly.

\begin{figure}[t]
\begin{center}
  \includegraphics[width=\linewidth]{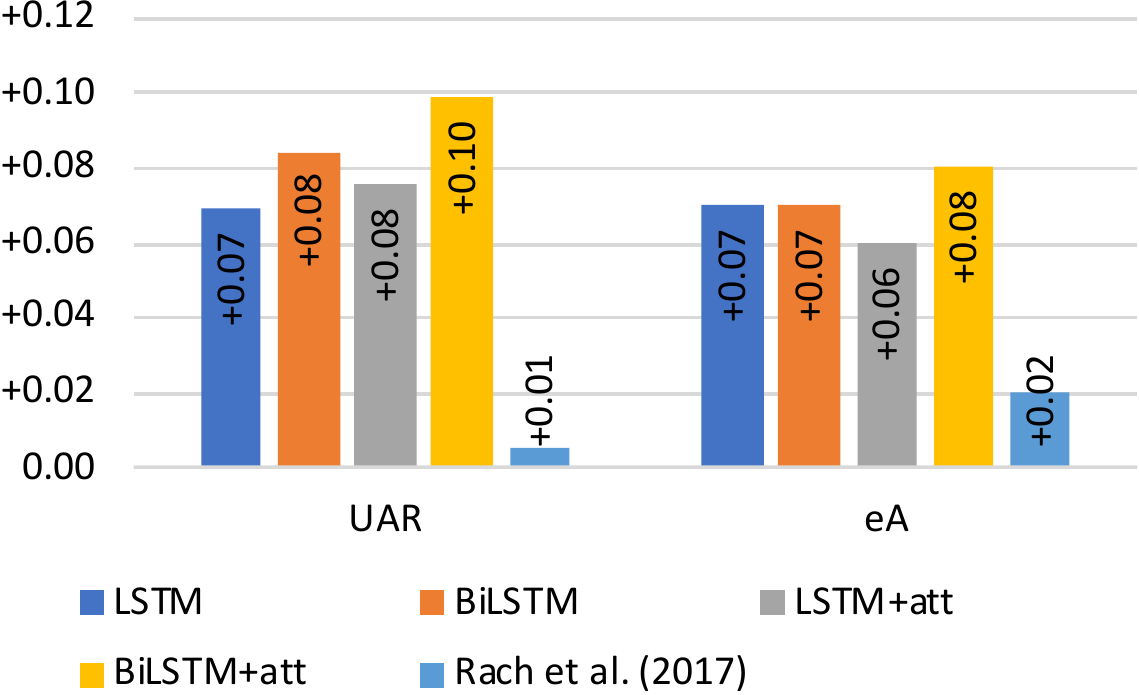}
  \caption{Absolute improvement of the IQ estimation models over the originally employed model by \cite{ultes2017domain} for IQ-based reward estimation with the dialogue-wise cross-validation setup. UAR and eA take values from 0 to 1}
  \label{fig:callimprovement}
\end{center}
\end{figure}

To alleviate this, a dialogue-wise cross-validation setup has been employed also consisting of 10 folds of disjoint sets of dialogues. By that, it can be guaranteed that there are no overlapping sub-dialogues in the training and test sets. All results of these experiments are presented in Table~\ref{tab:callxvalresults} with the absolute improvement of the two main measures UAR and eA over the SVM-based approach of \citet{ultes2015b} visualised in Figure~\ref{fig:callimprovement}.

The proposed BiLSTM+att model outperforms existing models and the baselines in all four performance measures by achieving an UAR of 0.54 and an eA of 0.94 after 40 epochs. Furthermore, both the BiLSTM and the attention mechanism by themselves improve the performance in terms of UAR. Based on this findings, the BiLSTM+att model is selected as reward estimator for the experiments in the dialogue policy learning setup as shown in Figure~\ref{fig:RLframework}.

\subsection{Dialogue Policy Learning}
% \todo[inline]{
% 
% - metrics for classification, dialogue
% -  
% 
% }

% \subsection{Experimental Setup}
% \todo[inline]{
% - algorithm: GP-SARSA cite milica
% - summary space, mapping
% - system actions to choose from
% - domains
% - belief tracker
% - evaluation measures: task success rate, interaction quality
% }

%\usepackage{graphics} is needed for \includegraphics
\begin{figure*}[t]
\begin{center}
  \includegraphics[width=0.95\linewidth]{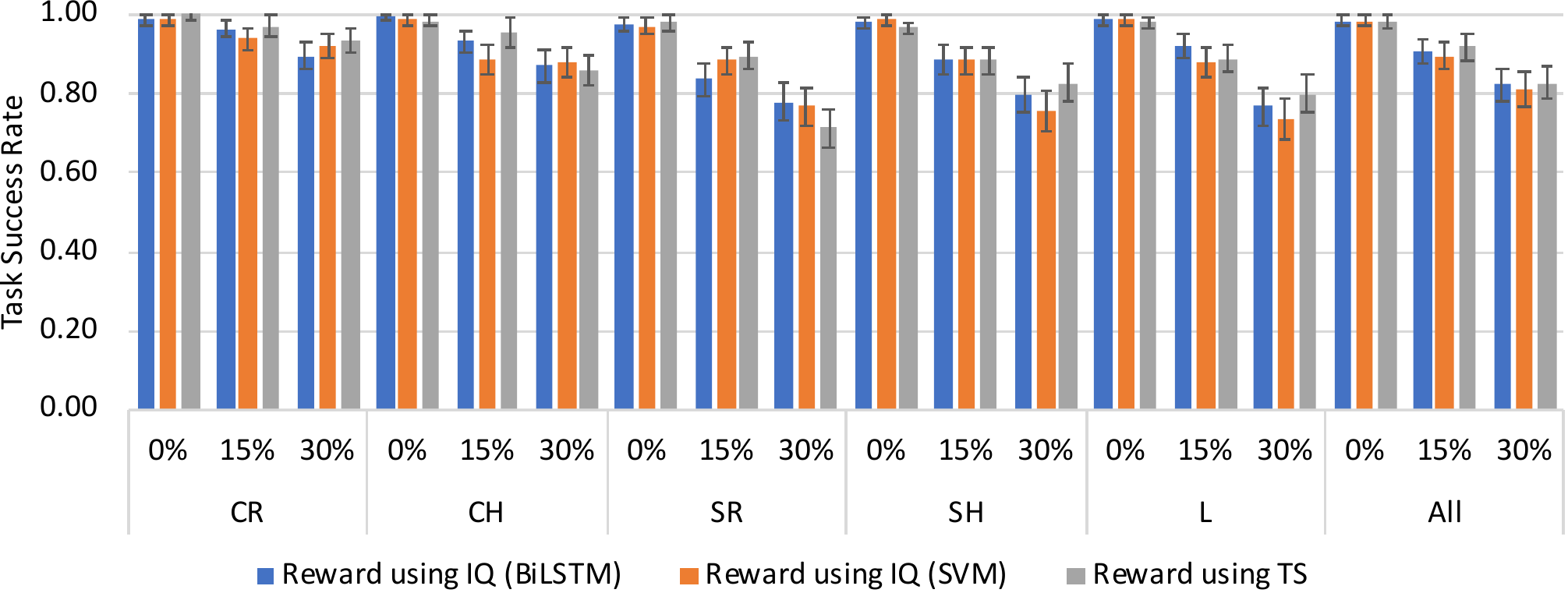}
  \caption{Results of the simulated experiments for all domains showing task success rate (TSR) only. Each value is computed after 100 evaluation / 1,000 training dialogues averaged over three trials. Numerical results with significance indicators are shown in Table~\ref{tab:results_simulation}.}
  \label{fig:result_simulation_TSR}
\end{center}
\end{figure*}

To analyse the impact of the IQ reward estimator on the resulting dialogue policy, experiments are conducted comparing three different reward models. The two baselines are in accordance to \citet{ultes2017domain}: having the objective task success as principal reward component ($R_{TS}$) and having the interaction quality estimated by a support vector machine as principal reward component ($R_{IQ}^{s}$). TS can be computed by comparing the outcome of each dialogue with the pre-defined goal. Of course, this is only possible in simulation and when evaluating with paid subjects. This goal information is not available to the IQ estimators, nor is it required. Both baselines are compared to our proposed BiLST model to estimate the interaction quality used as principal reward component ($R_{IQ}^{bi}$).

For learning the dialogue behaviour, a policy model based on the GP-SARSA algorithm~\cite{gasic2014gaussian} is used. This is a value-based method that uses a Gaussian process to approximate the state-value function. As it takes into account the uncertainty of the approximation, it is very sample efficient and may even be used to learn a policy directly through real human interaction~\cite{gasic2013}. 

The decisions of the policy are based on a summary space representation of the dialogue state tracker. In this work, the focus tracker~\cite{henderson2014second}---an effective rule-based tracker---is used. For each dialogue decision, the policy chooses exactly one summary action out of a set of summary actions which are based on general dialogue acts like \textit{request}, \textit{confirm} or \textit{inform}. The exact number of system actions varies for the domains and ranges from 16 to 25. %All components are implemented in the PyDial system~\cite{ultes2017pydial}.

% The IQ reward estimator is evaluated against the baseline of using the traditional reward function based on task success (TS). While IQ needs to be estimated, TS can be computed by comparing the outcome of each dialogue with the pre-defined goal.  Of course, this is only possible in simulation and when evaluating with paid subjects. This goal information is not available to the IQ estimator, nor is it required.

To measure the dialogue performance, the task success rate (TSR) and the average interaction quality (AIQ) are measured:
%(in correspondence with the two objective functions)
the TSR represents the ratio of dialogues for which the system was able to provide the correct result.
% (i.e. the user goal matches the information the system acquired during the interaction). 
AIQ is calculated based on the estimated IQ values of the respective model ($AIQ^{bi}$ for the BiLSTM and $AIQ^{s}$ for the SVM) at the end of each dialogue. As there are two IQ estimators, a distinction is made between $AIQ^{s}$ and $AIQ^{bi}$. Additionally, the average dialogue length (ADL) is reported. 
%\mgcomment{Have you tried using IQ as immediate reward?}

% Table generated by Excel2LaTeX from sheet 'table TS coop'
\begin{table}[t]
  \centering
  \footnotesize
  \caption{Statistics of the domains the IQ reward estimator is trained on (LetsGo) and applied to (rest).}
    \begin{tabular}{rccc}
    \toprule
    \textit{Domain} & \textit{Code} & \textit{\# constraints} & \textit{\# DB items} \\
    \midrule
    LetsGo &       & 4     & - \\
    \cmidrule{2-4}
    CamRestaurants & CR    & 3     & 110 \\
    CamHotels & CH    & 5     & 33 \\
    SFRestaurants & SR    & 6     & 271 \\
    SFHotels & SH    & 6     & 182 \\
    Laptops & L     & 6     & 126  \\
%    TV & TV     & 6     & 94 \\
    \bottomrule
    \end{tabular}%
  \label{tab:domains}%
  %\vspace{-6pt}
\end{table}%

For the simulation experiments, the performance of the trained polices on five different domains was evaluated: Cambridge Hotels and Restaurants, San Francisco Hotels and Restaurants, and Laptops. The complexity of each domain is shown in Table~\ref{tab:domains} and compared to the LetsGo domain (the domain the estimators have been trained on).

% Table generated by Excel2LaTeX from sheet 'table TS coop'
\begin{table*}[t]
  \centering
  \footnotesize
  \caption{Results of the simulated experiments for all domains showing task success rate (TSR), average interaction quality estimated with the SVM ($AIQ^{s}$) and the BiLSTM ($AIQ^{bi}$, and average dialogue length (ADL) in number of turns. Each value is computed after 100 evaluation / 1,000 training dialogues averaged over three trials with different random seeds. $^{1,2,3}$ marks statistically significant difference compared to $R_{TS}$, to $R_{IQ}^{s}$, and to {$AIQ^{bi}$}, respectively ($p<0.05$, T-test for TSR and ADL, Mann-Whitney-U test for AIQ).}
  \setlength{\tabcolsep}{3pt}
    \begin{tabular}{crcccccccccc}
    \toprule
    \multirow{2}[0]{*}{\textit{Domain}} & \multicolumn{1}{c}{\multirow{2}[0]{*}{\textit{SER}}} & \multicolumn{3}{c}{\textit{TSR}} & \multicolumn{2}{c}{$AIQ^{s}$} & \multicolumn{2}{c}{$AIQ^{bi}$} & \multicolumn{3}{c}{\textit{ADL}} \\
    \cmidrule(l{2pt}r{2pt}){3-5} \cmidrule(l{2pt}r{2pt}){6-7} \cmidrule(l{2pt}r{2pt}){8-9} \cmidrule(l{2pt}r{2pt}){10-12}
          &       & $R_{TS}$ & $R_{IQ}^{s}$ & $R_{IQ}^{bi}$ & $R_{TS}$ & $R_{IQ}^{s}$ & $R_{TS}$ & $R_{IQ}^{bi}$ & $R_{TS}$ & $R_{IQ}^{s}$ & $R_{IQ}^{bi}$ \\
          \midrule
    \multirow{3}[0]{*}{CR} 
    & 0\%   & \textbf{1.00}$^{2,3}$ & 0.99$^{1}$      & 0.99$^{1}$    & 3.64$^{2}$          & \textbf{3.90}$^{1}$ & 3.68$^{3}$          & \textbf{3.83}$^{1}$ & 4.68                & 4.88                & \textbf{4.59} \\
    & 15\%  & \textbf{0.97}         & 0.94            & 0.96          & 3.35$^{2}$          & \textbf{3.65}$^{1}$ & 3.45$^{3}$          & \textbf{3.63}$^{1}$ & 5.85$^{3}$          & 5.33                & \textbf{5.10}$^{1}$ \\
    & 30\%  & \textbf{0.94}         & 0.92            & 0.90          & 3.15$^{2}$          & \textbf{3.34}$^{1}$ & 3.22                & \textbf{3.30}       & 6.34                & 6.30                & \textbf{6.25} \\
          \cmidrule(l{3pt}r{3pt}){2-12}
    \multirow{3}[0]{*}{CH} 
    & 0\%   & 0.98                  & 0.99            & \textbf{0.99} & 3.26$^{2}$          & \textbf{3.62}$^{1}$ & 3.33                & \textbf{3.44}       & 5.71                & 5.61                & \textbf{5.40} \\
    & 15\%  & \textbf{0.96}$^{2}$   & 0.89$^{1,3}$    & 0.93$^{2}$    & \textbf{2.90}       & 2.88                & \textbf{3.14}       & 3.14                & \textbf{6.28}$^{2}$ & 7.26$^{1,3}$        & 6.31$^{2}$ \\
    & 30\%  & 0.86                  & \textbf{0.88}   & 0.87          & 2.38$^{2}$          & \textbf{2.79}$^{1}$ & 2.79$^{3}$          & \textbf{3.02}$^{1}$ & 7.94$^{3}$          & 7.31                & \textbf{6.99}$^{1}$ \\
          \cmidrule(l{3pt}r{3pt}){2-12}
    \multirow{3}[0]{*}{SR} 
    & 0\%   & \textbf{0.98}         & 0.97            & 0.98          & 3.04$^{2}$          & \textbf{3.53}$^{1}$ & 3.13$^{3}$          & \textbf{3.37}$^{1}$ & 6.26                & 6.03                & \textbf{5.80} \\
    & 15\%  & \textbf{0.90}$^{3}$   & 0.88            & 0.84$^{1}$    & 2.40$^{2}$          & \textbf{3.00}$^{1}$ & 2.85$^{3}$          & \textbf{3.01}$^{1}$ & 7.99                & 7.55                & \textbf{7.33} \\
    & 30\%  & 0.71                  & 0.77            & \textbf{0.78} & 2.03$^{2}$          & \textbf{2.52}$^{1}$ & 2.46$^{3}$          & \textbf{2.78}$^{1}$ & 9.77$^{3}$          & 9.41                & \textbf{8.50}$^{1}$ \\
          \cmidrule(l{3pt}r{3pt}){2-12}
    \multirow{3}[0]{*}{SH} 
    & 0\%   & 0.97                  & \textbf{0.99}   & 0.98          & 3.15$^{2}$          & \textbf{3.52}$^{1}$ & 3.17$^{3}$          & \textbf{3.36}$^{1}$ & 5.99$^{2}$          & \textbf{5.50}$^{1}$ & 5.76 \\
    & 15\%  & 0.88                  & 0.88            & \textbf{0.89} & 2.63$^{2}$          & \textbf{2.94}$^{1}$ & 2.77$^{3}$          & \textbf{3.17}$^{1}$ & 7.98$^{3}$          & 7.59$^{3}$          & \textbf{6.63}$^{1,2}$ \\
    & 30\%  & \textbf{0.83}$^{2}$   & 0.76$^{1}$      & 0.80          & 2.50                & \textbf{2.63}       & 2.70$^{3}$          & \textbf{2.87}$^{1}$ & 8.38                & 9.21                & \textbf{8.37} \\
          \cmidrule(l{3pt}r{3pt}){2-12}
    \multirow{3}[0]{*}{L} 
    & 0\%   & 0.98                  & \textbf{0.99}   & \textbf{0.99} & 3.26$^{2}$          & \textbf{3.61}$^{1}$ & 3.28                & \textbf{3.41}       & 5.78                & \textbf{5.44}       & 5.60 \\
    & 15\%  & 0.89                  & 0.88            & \textbf{0.92} & 2.58$^{2}$          & \textbf{2.97}$^{1}$ & 2.92$^{3}$          & \textbf{3.17}$^{1}$ & 7.19                & 7.34                & \textbf{6.73} \\
    & 30\%  & \textbf{0.80}         & 0.74            & 0.77          & 2.43                & \textbf{2.57}       & 2.79                & \textbf{2.92}       & 8.22$^{2}$          & 9.32$^{1,3}$        & \textbf{7.97}$^{2}$ \\
%           \cmidrule(l{3pt}r{3pt}){2-12}
%     \multirow{3}[0]{*}{TV} & 0\%   & \textbf{0.98} & 0.97  & 0.98  & 3.05$^{1}$  & \textbf{3.73}$^{1}$ & 3.24  & \textbf{3.45} & 6.17  & \textbf{5.57} & 5.69 \\
%           & 15\%  & \textbf{0.90} & 0.88  & 0.90  & 2.70$^{1}$  & \textbf{3.14}$^{1}$ & 2.99  & \textbf{3.10} & 7.48  & 7.29  & \textbf{7.04} \\
%           & 30\%  & 0.82  & \textbf{0.83} & 0.79  & 2.47$^{1}$  & \textbf{2.95}$^{1}$ & 2.72  & \textbf{2.93} & 8.57  & \textbf{7.85} & 7.87 \\
          \cmidrule(l{3pt}r{3pt}){2-12}
    \multirow{3}[0]{*}{All} 
    & 0\%   & 0.98                  & 0.98            & \textbf{0.98} & 3.23$^{2}$          & \textbf{3.65}$^{1}$ & 3.31                & \textbf{3.48}       & 5.76                & 5.50                & \textbf{5.47} \\
    & 15\%  & \textbf{0.92}         & 0.89            & 0.91          & 2.76$^{2}$          & \textbf{3.10}$^{1}$ & 3.02$^{2}$0          & \textbf{3.20}$^{1}$ & 7.13                & 7.06                & \textbf{6.52} \\
    & 30\%  & \textbf{0.83}         & 0.81            & 0.82          & 2.49                & \textbf{2.80}       & 2.78                & \textbf{2.97}       & 8.20$^{2}$          & 8.23$^{1,3}$        & \textbf{7.66}$^{2}$ \\
          \bottomrule
    \end{tabular}%
  \label{tab:results_simulation}%
\end{table*}%

The dialogues were created using the publicly available spoken dialogue system toolkit PyDial~\cite{ultes2017pydial}\footnote{Code freely available at \url{http://www.pydial.org}} which contains an implementation of the agenda-based user simulator~\cite{schatzmann2009} with an additional error model. The error model simulates the required semantic error rate (SER) caused in the real system by the noisy speech channel. For each domain, all three reward models are compared on three SERs: 0\%, 15\%, and 30\%. More specifically, the applied evaluation environments are based on Env.~1, Env.~3, and Env.~6, respectively, as defined by \citet{casanueva2017benchmarking}. Hence, for each domain and for each SER, policies have been trained using 1,000 dialogues followed by an evaluation step of 100 dialogues. The task success rates in Figure~\ref{fig:result_simulation_TSR} with exact numbers shown in 
Table~\ref{tab:results_simulation} were computed based on the evaluation step averaged over three train/evaluation cycles with different random seeds.

%$^{1}$
As already known from the experiments conducted by \citet{ultes2017domain}, the results of the SVM IQ reward estimator show similar results in terms of TSR for $R_{IQ}^{s}$ and $R_{TS}$ in all domains for an SER of 0\%. This finding is even stronger when comparing $R_{IQ}^{bi}$ and $R_{TS}$. These high TSRs are achieved while having the dialogues of both IQ-based models result in higher AIQ values compared to $R_{TS}$ throughout the experiments. Of course, only the  IQ-based model is aware of the IQ concept and indeed is trained to optimise it.

For higher SERs, the TSRs lightly degrade for the IQ-based reward estimators. However, there seems to be a tendency that the TSR for $R_{IQ}^{bi}$ is more robust against noise compared to $R_{IQ}^{s}$ while still resulting in better AIQ values.

Finally, even though the differences are mostly not significant, there is also a tendency for $R_{IQ}^{bi}$ to result in shorter dialogues compared to both $R_{IQ}^{s}$ and $R_{TS}$.

\section{Discussion}
One of the major questions of this work addresses the impact of an IQ reward estimator on the resulting dialogues where the IQ estimator achieves better performance than previous ones. Analysing the results of the dialogue policy learning experiment leads to the conclusion that the policy learned with $R_{IQ}^{bi}$ performs similar or better than $R_{IQ}^{s}$ through out all experiments while still achieving better average user satisfaction compared to $R_{TS}$. Especially for noisy environments, the improvement is relevant. 

The BiLSTM clearly performs better on the LEGO corpus while learning the temporal dependencies instead of using handcrafted ones. However, it entails the risk that these learned temporal dependencies are too specific to the original data so that the model does not generalise well anymore. This would mean that it would be less suitable to be applied to dialogue policy learning for different domains. Luckily, the experiments clearly show that this is not the case. 

%one fear we had: as the temporal dependencies are learned and are not handcrafted anymore, they might end up to be too specialized for learning policies in other domains. this has shown to not be a problem at all.

Obviously, the experiments have only been conducted in a simulated environment and not verified in a user study with real humans. However, the general framework of applying an IQ reward estimator for learning a dialogue policy has already been successfully validated with real user experiments by ~\citet{ultes2017domain} and it seems rather unlikely that the changes we induce by changing the reward estimator lead to a fundamentally different result.

% not evaluated with real users, however, original contribution already successfully validated whole setup with real user experumetns.
% 

\section{Conclusion}
In this work we proposed a novel model for interaction quality estimation based on BiLSTMs with attention mechanism that clearly outperformed the baseline while learning all temporal dependencies implicitly. Furthermore, we analysed the impact of the performance increase on learned polices that use this interaction quality estimator as the principal reward component. The dialogues of the proposed interaction quality estimator show a slightly higher robustness towards noise and shorter dialogues while still yielding good performance in terms of both of task  success  rate  and  (estimated)  user  satisfaction. This  has been demonstrated by training the reward estimator on a bus information domain and applying it to learn dialogue policies in five different domains (Cambridge restaurants and hotels, San Francisco restaurants and hotels, Laptops) in a simulated experiment.

For future work, we aim at extending the interaction quality estimator by incorporating domain-independent linguistic data to further improve the estimation performance. Furthermore, the effects of using a user satisfaction-based reward estimator needs to be applied to more complex tasks.%, e.g., as defined by the Conversational Entitiy Dialogue Model~\cite{ultes2018addressing}.

\bibliography{references}
\bibliographystyle{acl_natbib}

\appendix

\end{document}